\documentclass[letterpaper, 10 pt, conference]{ieeeconf}

\IEEEoverridecommandlockouts
\usepackage[T1]{fontenc}
\usepackage{times}
\usepackage{amsmath}
\usepackage{amssymb}
\usepackage{graphicx}
\usepackage[ruled,linesnumbered]{algorithm2e}
\usepackage{booktabs}
\usepackage{float}
\usepackage{bigstrut}
\usepackage{multirow}
\usepackage{balance}
\usepackage[hidelinks=true,bookmarks=FWLse]{hyperref}
\usepackage{caption}
\usepackage{lineno}
\usepackage{amsopn}
\usepackage{comment}
\usepackage{color}
\usepackage{cite}
\usepackage{algorithmic}
\usepackage{tabularx}
\usepackage{xcolor}
\usepackage{mathrsfs}
\usepackage{url,subfigure,tabulary}
\usepackage[cmintegrals]{newtxmath}

\DeclareMathOperator*{\argmin}{arg\,min}

\usepackage{array}
\makeatletter
\newcommand{\thickhline}{%
    \noalign {\ifnum 0=`}\fi \hrule height 1pt
    \futurelet \reserved@a \@xhline
}
\newcolumntype{"}{@{\hskip\tabcolsep\vrule width 1pt\hskip\tabcolsep}}
\makeatother

\makeatletter

\newcommand{\Rmnum}[1]{\expandafter\@slowromancap\romannumeral #1@}
\makeatother

\newcolumntype{L}[1]{>{\raggedright\arraybackslash}p{#1}}
\newcolumntype{C}[1]{>{\centering\arraybackslash}p{#1}}
\newcolumntype{R}[1]{>{\raggedleft\arraybackslash}p{#1}}

\title{\LARGE \bf
Applying Surface Normal Information in Drivable Area and Road Anomaly Detection for Ground Mobile Robots
}

\author{Hengli Wang$^1$*, Rui Fan$^2$*, Yuxiang Sun$^1$, and Ming Liu$^1$, \IEEEmembership{Senior Member, IEEE}
\thanks{This work was supported by the National Natural Science Foundation of China, under grant No. U1713211, Collaborative Research Fund by Research Grants Council Hong Kong, under Project No. C4063-18G, and the Research Grant Council of Hong Kong SAR Government, China, under Project No. 11210017, awarded to Prof. Ming Liu. \textit{(Corresponding author: Ming Liu.)}}
\thanks{$^1$H. Wang, Y. Sun and M. Liu are with the Department of Electronic and Computer Engineering, the Hong Kong University of Science and Technology, Clear Water Bay, Kowloon, Hong Kong SAR, China (email: \{hwangdf, eeyxsun, eelium\}@ust.hk).}
\thanks{$^2$R. Fan is with the Jacobs School of Engineering as well as the UCSD Health, the University of California, San Diego, La  Jolla, CA 92093, U.S. (email: rui.fan@ieee.org). }
\thanks{*The authors contributed equally to this work.}
}

\begin{document}

\maketitle
\thispagestyle{empty}
\pagestyle{empty}

\begin{abstract}
    The joint detection of drivable areas and road anomalies is a crucial task for ground mobile robots. In recent years, many impressive semantic segmentation networks, which can be used for pixel-level drivable area and road anomaly detection, have been developed. However, the detection accuracy still needs improvement. Therefore, we develop a novel module named the Normal Inference Module (NIM), which can generate surface normal information from dense depth images with high accuracy and efficiency. Our NIM can be deployed in existing convolutional neural networks (CNNs) to refine the segmentation performance. To evaluate the effectiveness and robustness of our NIM, we embed it in twelve state-of-the-art CNNs. The experimental results illustrate that our NIM can greatly improve the performance of the CNNs for drivable area and road anomaly detection. Furthermore, our proposed NIM-RTFNet ranks 8th on the KITTI road benchmark and exhibits a real-time inference speed.
\end{abstract}

\section{Introduction}
Ground mobile robots, such as sweeping robots and robotic wheelchairs, are playing significant roles in improving the quality of human life \cite{wang2019self,sun2017improving,sun2018motion}. Visual environment perception and autonomous navigation are two fundamental components for ground mobile robots. The former takes as input sensory data and outputs environmental perception results, with which the latter automatically moves the robot from point A to point B. Among the environment perception tasks for ground mobile robots, the joint detection of drivable areas and road anomalies is a critical component that labels the image as the drivable area or road anomaly at the pixel-level. In this paper, the drivable area refers to a region where ground mobile robots can pass through, while a road anomaly refers to an area with a large difference in height from the surface of the drivable area. Accurate and real-time drivable area and road anomaly detection could avoid accidents for ground mobile robots.

With the great advancement of deep learning technologies, many effective semantic segmentation networks that could be used for the task of drivable area and road anomaly detection have been proposed \cite{chen2018encoder,wang2018depth}. Specifically, Chen \textit{et al.} \cite{chen2018encoder} proposed DeepLabv3+, which combines the spatial pyramid pooling (SPP) module and the encoder-decoder architecture to generate accurate semantic predictions. However, most of the networks simply use RGB images, which suffer from degraded illumination conditions \cite{fan2019pothole}. Recently, some data-fusion convolutional neural networks (CNNs) have been proposed to improve the accuracy of semantic segmentation. Such architectures generally utilize two different types of sensory data to learn informative learning representations. For example, Wang \textit{et al.} \cite{wang2018depth} proposed a novel depth-aware CNN to fuse depth images with RGB images, which has improved the performance of semantic segmentation. Thus, the fusion of different modalities of data is a promising research direction that deserves more attention.

\begin{figure}[t]
    \centering
    \includegraphics[width=\linewidth]{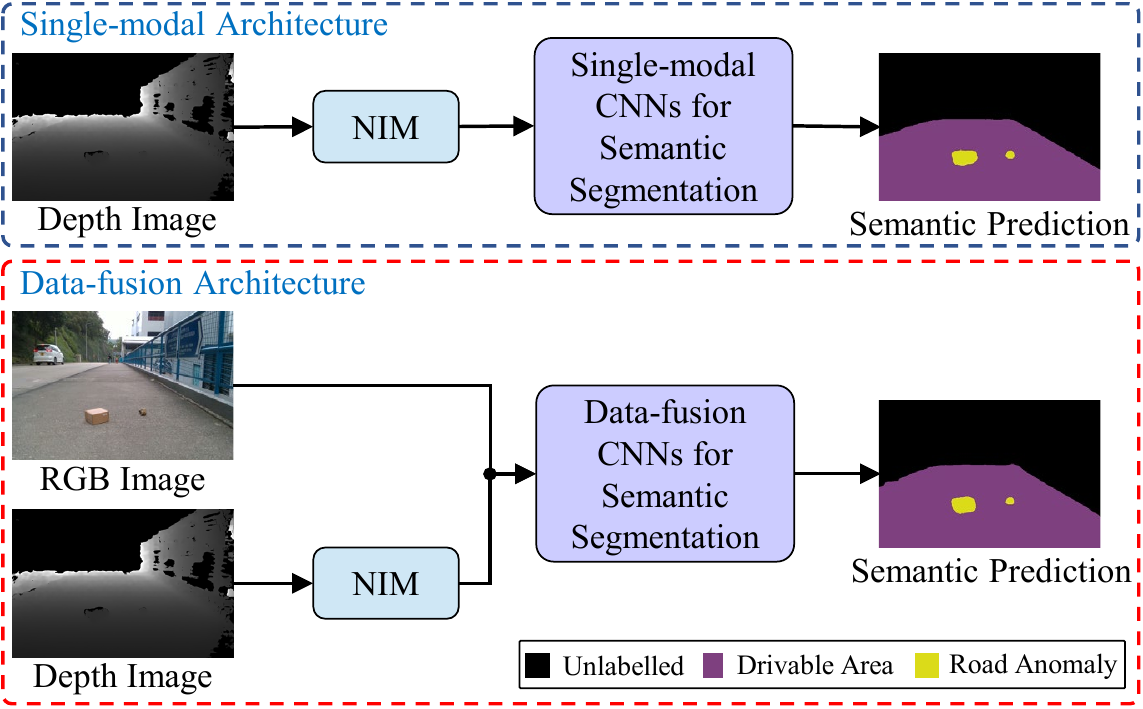}
    \caption{The overview of our proposed CNN architecture for detecting drivable areas and road anomalies, where our proposed NIM can be deployed in existing single-modal or data-fusion CNNs to refine the segmentation performance.}
    \label{fig.framework}
    \vspace{-1em}
\end{figure}

In this paper, we first introduce a novel module named Normal Inference Module (NIM), which generates surface normal information from dense depth images with high accuracy and efficiency. The surface normal information serves as a different modality of data, which can be deployed in existing semantic segmentation networks to improve performance, as illustrated in Fig. \ref{fig.framework}. To validate the effectiveness and robustness of our proposed NIM, we use our previous ground mobile robots perception (GMRP) dataset\footnote{\url{https://github.com/hlwang1124/GMRPD}} \cite{wang2019self} to train twelve state-of-the-art CNNs (eight single-modal CNNs and four data-fusion CNNs) with and also without our proposed NIM embedded. The experimental results demonstrate that our proposed NIM can greatly enhance the performance of the aforementioned CNNs for the task of drivable area and road anomaly detection. Furthermore, our proposed NIM-RTFNet ranks 8th on the KITTI road benchmark\footnote{\url{www.cvlibs.net/datasets/kitti/eval_road.php}} \cite{Fritsch2013ITSC} and exhibits a real-time inference speed. The contributions of this paper are summarized as follows:

\begin{itemize}
    \item We develop a novel NIM and show its effectiveness on improving the semantic segmentation performance.
    \item We conduct extensive studies on the impact of different modalities of data on semantic segmentation networks.
    \item Our proposed NIM-RTFNet greatly minimizes the trade-off between speed and accuracy on the KITTI road benchmark.
\end{itemize}

\begin{figure*}[!t]
    \centering
    \includegraphics[width=0.8\textwidth]{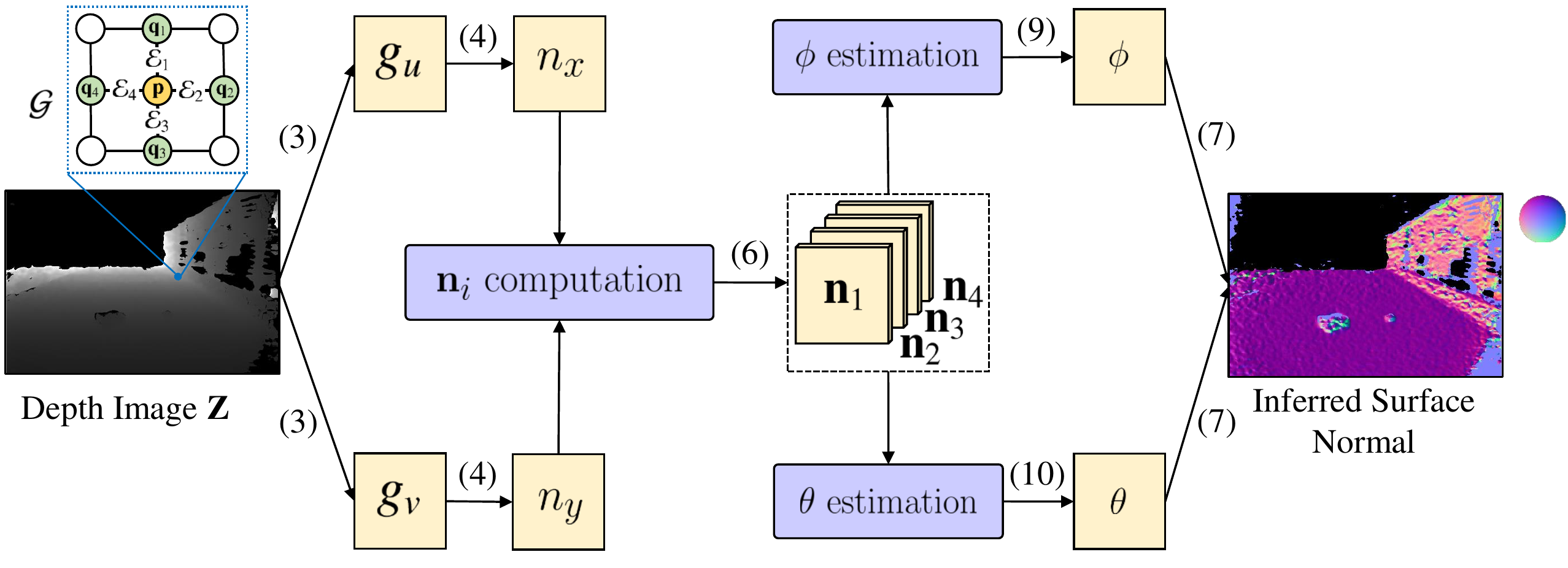}
    \caption{Illustration of our proposed NIM, where the numbers within parentheses denote the corresponding equations. We first use two kernels to compute volumes, and then solve an optimization problem to generate surface normal images.}
    \label{fig.nim}
\end{figure*}

\section{Related Work}
In this section, we briefly overview twelve state-of-the-art semantic segmentation networks, including eight single-modal networks, \textit{i.e.,} fully convolutional network (FCN) \cite{long2015fully}, SegNet \cite{badrinarayanan2017segnet}, U-Net \cite{ronneberger2015u}, DeepLabv3+ \cite{chen2018encoder}, DenseASPP \cite{yang2018denseaspp}, DUpsampling \cite{tian2019decoders}, ESPNet \cite{mehta2018espnet} and Gated-SCNN (GSCNN) \cite{takikawa2019gated}, as well as four data-fusion networks, \textit{i.e.,} FuseNet \cite{hazirbas2016fusenet}, Depth-aware CNN \cite{wang2018depth}, MFNet \cite{ha2017mfnet} and RTFNet \cite{sun2019rtfnet}.

\subsection{Single-modal CNN Architectures}
FCN \cite{long2015fully} was the first end-to-end semantic segmentation network. Of the three FCN variants, FCN-32s, FCN-16s and FCN-8s, we use FCN-8s in our experiments. SegNet \cite{badrinarayanan2017segnet} first presented the encoder-decoder architecture, which is widely used in current networks. U-Net \cite{ronneberger2015u} was designed based on an FCN \cite{long2015fully}, and adds skip connections between the encoder and decoder to improve the information flow.

DeepLabv3+ \cite{chen2018encoder} was designed to combine the advantages of both the SPP module and the encoder-decoder architecture. To make the feature resolution sufficiently dense for autonomous driving, DenseASPP \cite{yang2018denseaspp} was proposed to connect a set of atrous convolutional layers in a dense way.

Different from the networks mentioned above, DUpsampling \cite{tian2019decoders} adopts a data-dependent decoder, which exploits the redundancy in the label space of semantic segmentation and has the ability to recover the pixel-wise prediction from low-resolution outputs of networks. ESPNet \cite{mehta2018espnet} employs a novel convolutional module named efficient spatial pyramid (ESP) to save computation and memory cost. GSCNN \cite{takikawa2019gated} utilizes a novel architecture consisting of a shape branch and a regular branch to focus on the boundary information.

\subsection{Data-fusion CNN Architectures}
FuseNet \cite{hazirbas2016fusenet} was proposed for the problem of semantic image segmentation using RGB-D data. It employs the popular encoder-decoder architecture, and adopts element-wise summation to combine the feature maps of the RGB stream and the depth stream. Depth-aware CNN \cite{wang2018depth} introduces two novel operations: depth-aware convolution and depth-aware average pooling, and leverages depth similarity between pixels to incorporate geometric information into the CNN.

MFNet \cite{ha2017mfnet} was proposed for semantic image segmentation using RGB-thermal images. It focuses on retaining the segmentation accuracy during real-time operation. RTFNet \cite{sun2019rtfnet} was developed to enhance the performance of semantic image segmentation using RGB-thermal images. The key component of RTFNet is the novel decoder, which includes short-cuts to keep more detailed information.

\section{Methodology}
\label{sec.method}
Our proposed NIM, as illustrated in Fig. \ref{fig.nim}, can generate surface normal information from dense depth images with both high precision and efficiency. The most common way of estimating the surface normal $\mathbf{n}=[n_x,n_y,n_z]^\top$ of a given 3D point $\mathbf{p}^\text{C}=[x,y,z]^\top$ in the camera coordinate system (CCS) is to fit a local plane: $\mathbf{n}^\top\mathbf{p}^\text{C}+\beta=0$ to $\mathbf{Q}=[\mathbf{p}^\text{C},\mathbf{q}_1,\dots,\mathbf{q}_k]^\top$, where $\mathbf{q}_1,\dots,\mathbf{q}_k$ are a collection of $k$ nearest neighboring points of $\mathbf{p}^\text{C}$.
For a pinhole camera model, $\mathbf{p}^\text{C}$ is linked with a pixel ${\mathbf{p}}^\text{I}=[u,v]^\top$ in the depth image $\mathbf{Z}$ by \cite{fan2018real}:
\begin{equation}
z\begin{bmatrix}
{\mathbf{p}}^\text{I}\\1
\end{bmatrix}=\begin{bmatrix}
f_x & 0 & u_\text{o}\\
0 & f_y & v_\text{o}\\
0 & 0 & 1
\end{bmatrix}\mathbf{p}^\text{C}.
\label{eq.intrinisc_matrix}
\end{equation}
A depth image can be considered as an undirected graph $\mathcal{G}=(\mathcal{P},\mathcal{E})$ \cite{fan2018road}, where $\mathcal{P}=\{\mathbf{p}^\text{I}_{11}, \mathbf{p}^\text{I}_{12}, \dots,\mathbf{p}^\text{I}_{mn}\}$ is a set of nodes (vertices) connected by edges $\mathcal{E}=\{(\mathbf{p}^\text{I}_{ij}, \mathbf{p}^\text{I}_{st})\ |\ \mathbf{p}^\text{I}_{ij}, \mathbf{p}^\text{I}_{st}\in\mathcal{P}\}$. Please note that in this paper, $\mathbf{p}^\text{I}_{ij}$ are simply written as $\mathbf{p}^\text{I}$. Plugging (\ref{eq.intrinisc_matrix}) into the local plane equation obtains \cite{fan2020three}:
\begin{equation}
\frac{1}{z}=-\frac{1}{\beta}\bigg(n_x\frac{u-u_\text{o}}{f_x}+n_y\frac{v-v_\text{o}}{f_y}+n_z\bigg).
\label{eq.1/z_n}
\end{equation}
Differentiating (\ref{eq.1/z_n}) with respect to $u$ and $v$ leads to
\begin{equation}
\begin{split}
\frac{\partial 1/z}{\partial u}=-\frac{1}{\beta f_x}n_x\approx\frac{1}{{\mathbf{Z}}(\mathbf{p}^\text{I}+[1,0]^\top)}-\frac{1}{{\mathbf{Z}}(\mathbf{p}^\text{I}-[1,0]^\top)}=g_u, \\ \frac{\partial 1/z}{\partial v}=-\frac{1}{\beta f_y}n_y\approx\frac{1}{{\mathbf{Z}}(\mathbf{p}^\text{I}+[0,1]^\top)}-\frac{1}{{\mathbf{Z}}(\mathbf{p}^\text{I}-[0,1]^\top)}=g_v.
\end{split}
\label{eq.d1/z_to_du_dv}
\end{equation}
Rearranging (\ref{eq.d1/z_to_du_dv}) results in
\begin{equation}
\begin{split}
n_x\approx-\beta f_x g_u, \ \ \  n_y\approx-\beta f_y g_v.
\end{split}
\label{eq.nx1_ny1}
\end{equation}
Given a pair of ${\mathbf{q}_{i}}$ and ${\mathbf{p}^\text{C}}$, we can work out the corresponding ${n_z}_i$ as follows:
\begin{equation}
{n_z}_i=\beta\Bigg(f_x\frac{\partial 1/z}{\partial u}\frac{\Delta {x_i}}{\Delta {z_i}} + f_y \frac{\partial 1/z}{\partial v} \frac{\Delta {y_i}}{\Delta {z_i}}    \Bigg),
\label{eq.nz1}
\end{equation}
where ${\mathbf{q}_i}-\mathbf{p}^\text{C}=[\Delta {x_i}, \Delta {y_i}, \Delta {z_i}]^\top$. Therefore, each neighboring point of $\mathbf{p}^\text{C}$ can produce a surface normal candidate as follows \cite{fan2020sne-roadseg}:
\begin{equation}
\mathbf{n}_{i}=\begin{bmatrix}
&-f_x \frac{\partial 1/z}{\partial u} \\
&-f_y \frac{\partial 1/z}{\partial v} \\
&f_x\frac{\partial 1/z}{\partial u}\frac{\Delta {x_i}}{\Delta {z_i}} + f_y \frac{\partial 1/z}{\partial v} \frac{\Delta {y_i}}{\Delta {z_i}}    \\
\end{bmatrix}.
\label{eq.nx2_ny2_nz2}
\end{equation}
The optimal surface normal
\begin{equation}
\hat{\mathbf{n}} = \begin{bmatrix}
\sin\theta\cos\phi\\
\sin\theta\sin\phi\\
\cos\theta
\end{bmatrix}
\label{eq.spherical_coordinates}
\end{equation}
can, therefore, be determined by finding the position at which the projections of $\bar{\mathbf{n}}_{i}=\frac{{\mathbf{n}}_{i}}{||{\mathbf{n}}_{i}||_2}=[\bar{n}_{x_i},\bar{n}_{y_i},\bar{n}_{z_i}]^\top$ distribute most intensively \cite{fan2019pothole}.
The visual perception module in a ground mobile robot should typically perform in real time, and taking more candidates into consideration usually makes the inference of $\hat{\mathbf{n}}$ more time-consuming. Therefore, we only consider the four neighbors adjacent to $\mathbf{p}^\text{I}$ in this paper.
$\hat{\mathbf{n}}$ can be estimated by solving \cite{fan2019pothole}
\begin{equation}
\underset{\phi,\theta}{\argmin}\sum_{i=1}^{4}-\hat{\mathbf{n}}\cdot{\bar{\mathbf{{n}}}}_{i},
\label{eq.energy_minimization}
\end{equation}
which has a closed-form solution as follows:
\begin{equation}
\phi=\arctan\bigg(\frac{\sum_{i=1}^{4}\bar{n}_{y_i}}{\sum_{i=1}^{4}\bar{n}_{x_i}}\bigg),
\label{eq.phi}
\end{equation}
\begin{equation}
\begin{split}
\theta= \arctan\Bigg(
\frac{1}{\sum_{i=1}^{4}\bar{n}_{z_i}}
\bigg(\sum_{i=1}^{4}\bar{n}_{x_i}\cos\phi+\sum_{i=1}^{4}\bar{n}_{y_i}\sin\phi\bigg)
\Bigg).
\end{split}
\label{eq.theta}
\end{equation}
Substituting (\ref{eq.theta}) and (\ref{eq.phi}) into (\ref{eq.spherical_coordinates}) results in the optimal surface normal inference, as shown in Fig. \ref{fig.nim}.

\begin{figure*}[t]
    \centering
    \includegraphics[width=\textwidth]{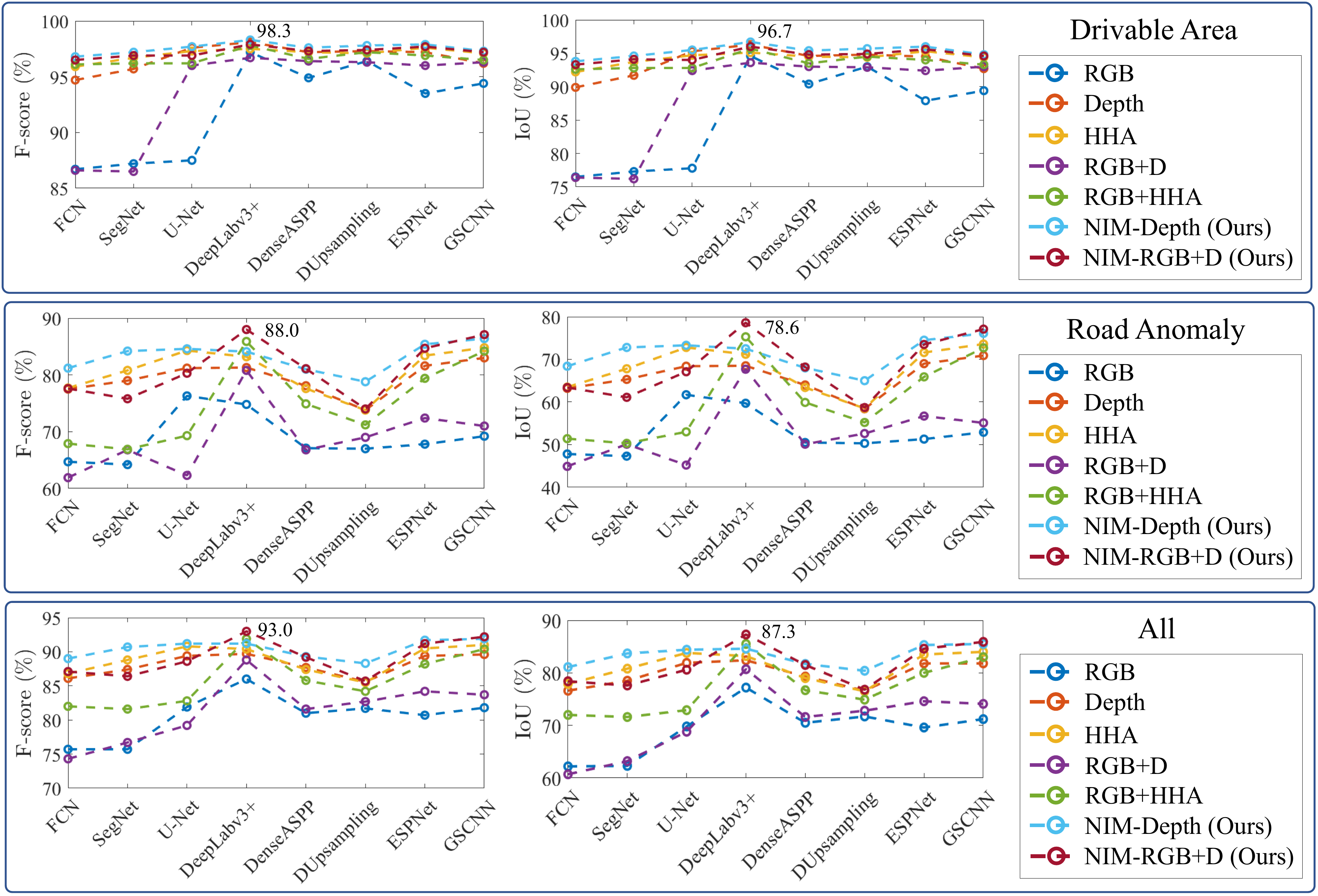}
    \caption{Performance comparison among the eight single-modal CNNs with seven setups on our GMRP dataset. The best result is highlighted in each subfigure.}
    \label{fig.one-stream}
\end{figure*}

\begin{figure*}[t]
    \centering
    \includegraphics[width=\textwidth]{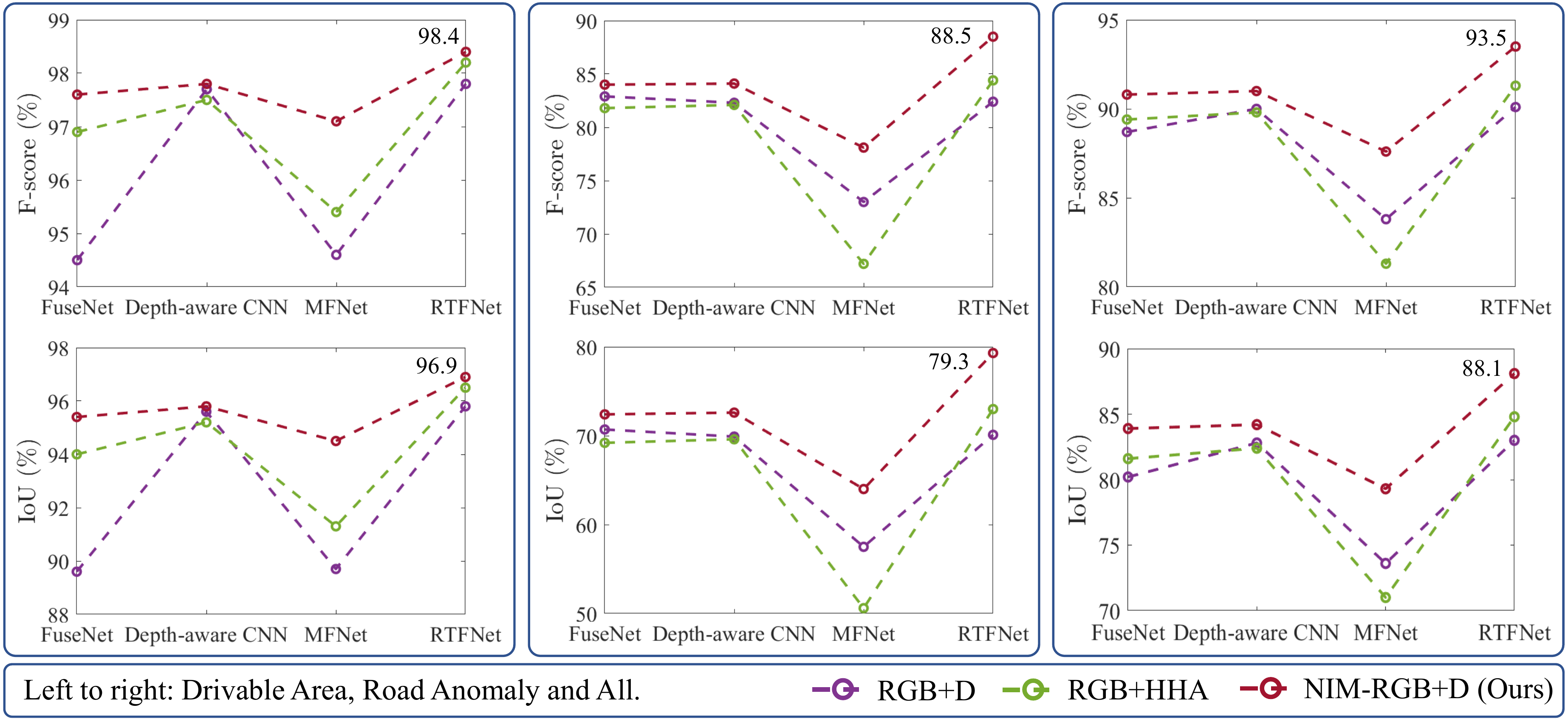}
    \caption{Performance comparison among the four data-fusion CNNs with three setups on our GMRP dataset. The best result is highlighted in each subfigure.}
    \label{fig.two-stream}
\end{figure*}

\begin{figure*}[t]
    \centering
    \includegraphics[width=\textwidth]{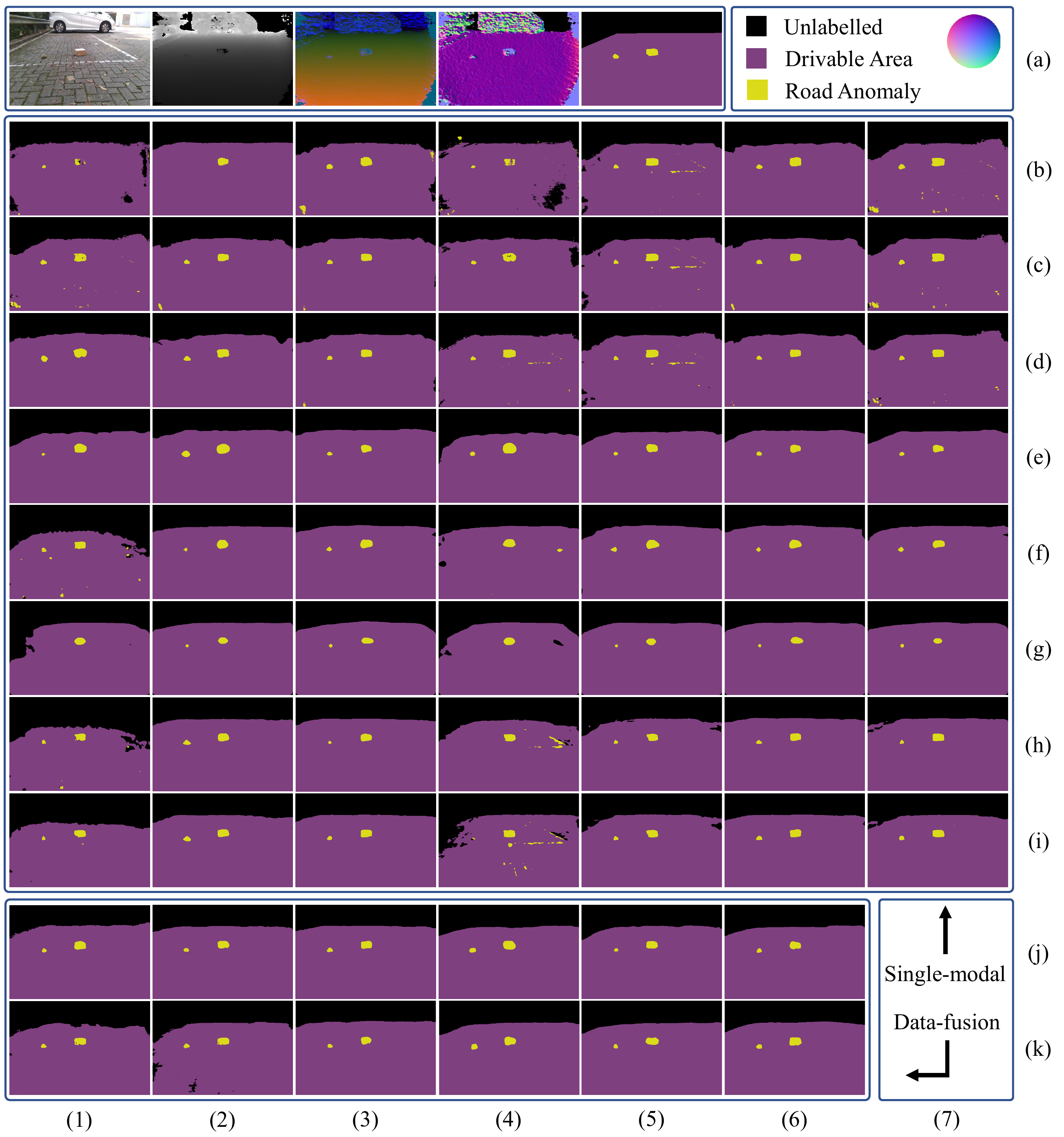}
    \caption{Examples of the experimental results on our GMRP dataset: columns (1)-(5) on row (a) show the RGB image, depth image, HHA image, surface normal image generated by our proposed NIM, and ground truth for semantic image segmentation, respectively; columns (1)-(7) on rows (b)-(i) show the semantic predictions obtained by seven setups on each of eight single-modal CNNs, which are respectively (1) RGB, (2) Depth, (3) HHA, (4) RGB+D, (5) RGB+HHA, (6) NIM-Depth (Ours), and (7) NIM-RGB+D (Ours), and (b) FCN, (c) SegNet, (d) U-Net, (e) DeepLabv3+, (f) DenseASPP, (g) DUpsampling, (h) ESPNet and (i) GSCNN; columns (1)-(3) on rows (j) and (k) show the semantic predictions obtained by three setups on two data-fusion CNNs, which are respectively (1) RGB+D, (2) RGB+HHA, and (3) NIM-RGB+D (Ours), and (j) FuseNet and (k) MFNet; columns (4)-(6) on rows (j) and (k) show the semantic predictions obtained by three setups on another two data-fusion CNNs, which are respectively (4) RGB+D, (5) RGB+HHA and (6) NIM-RGB+D (Ours), and (j) Depth-aware CNN and (k) RTFNet. The top right is the reference for surface normal images.}
    \label{fig.mainresult}
\end{figure*}

\section{Experimental Results and Discussions}

\subsection{Dataset Preparation and Experimental Setup}
We recently published a pixel-level drivable area and road anomaly detection dataset for ground mobile robots, named the GMRP dataset \cite{wang2019self}. Different from existing datasets, such as KITTI \cite{Fritsch2013ITSC} and Cityscapes \cite{Cordts2016Cityscapes}, our GMRP dataset covers the scenes and road anomalies that are common for ground mobile robots, \textit{e.g.,} sweeping robots and robotic wheelchairs. We refer readers to our previous paper \cite{wang2019self} for the details of the dataset.

In order to evaluate the effectiveness and robustness of our proposed NIM, we use our GMRP dataset to train twelve CNNs as mentioned above, including eight single-modal CNNs and four data-fusion CNNs. We train each single-modal CNN with seven setups. Specifically, we first train each one with input RGB, depth and HHA images (denoted as \textbf{RGB}, \textbf{Depth} and \textbf{HHA}), separately, where HHA \cite{hazirbas2016fusenet} is a three-channel feature map computed from the depth. Then, we train each one with input four-channel RGB-Depth and six-channel RGB-HHA (denoted as \textbf{RGB+D} and \textbf{RGB+HHA}), separately. Finally, we embed our proposed NIM in each single-modal CNN and train it with input depth images and four-channel RGB-Depth (denoted as \textbf{NIM-Depth} and \textbf{NIM-RGB+D}), separately. Similarly, we train each data-fusion CNN with three setups, separately denoted as \textbf{RGB+D}, \textbf{RGB+HHA} and \textbf{NIM-RGB+D}.

The total 3896 images in our GMRP dataset are split into a training set, a validation set and a testing set that contains 2726, 585 and 585 images, respectively. We train each network until the loss convergence and then select the best model according to the performance of the validation set. We adopt two metrics for the quantitative evaluations, the F-score and the Intersection over Union (IoU) for each class. We also compute the average values across two classes for the F-score and the IoU. The experimental results are presented in Section \ref{sec.ourdataset}.

To validate the effectiveness and robustness of our proposed NIM for autonomous cars, we also conduct experiments on the KITTI dataset. Since we focus on the detection of drivable areas and road anomalies, our task does not match the KITTI semantic image segmentation benchmark. However, our drivable area detection task perfectly matches the KITTI road benchmark \cite{Fritsch2013ITSC}. Therefore, we submit our best approach to the KITTI road benchmark. The experimental results are presented in Section \ref{sec.kittiroad}.

\subsection{Evaluations on Our GMRP Dataset}
\label{sec.ourdataset}
The performances of the single-modal and data-fusion CNNs mentioned above are compared in Fig. \ref{fig.one-stream} and Fig. \ref{fig.two-stream}, respectively. We can observe that the CNNs with our proposed NIM embedded (\textbf{NIM-Depth} or \textbf{NIM-RGB+D}) outperform those without NIM embedded. Fig. \ref{fig.mainresult} presents the sample qualitative results, where we can see that our proposed NIM greatly reduces the noise in the semantic predictions, especially for road anomaly detection.
% which is also confirmed by the qualitative results presented in Fig. \ref{fig.mainresult}.
Specifically, for the networks with \textbf{Depth} and \textbf{RGB+D} setup, embedding our proposed NIM increases the average F-score and IoU by around 3.3-12.8$\%$ and 5.1-17.7$\%$, respectively. Furthermore, RTFNet \cite{sun2019rtfnet} achieves the best overall performance.

\subsection{Evaluations on the KITTI Road Benchmark}
\label{sec.kittiroad}
\begin{figure}[!t]
    \centering
    \includegraphics[width=\linewidth]{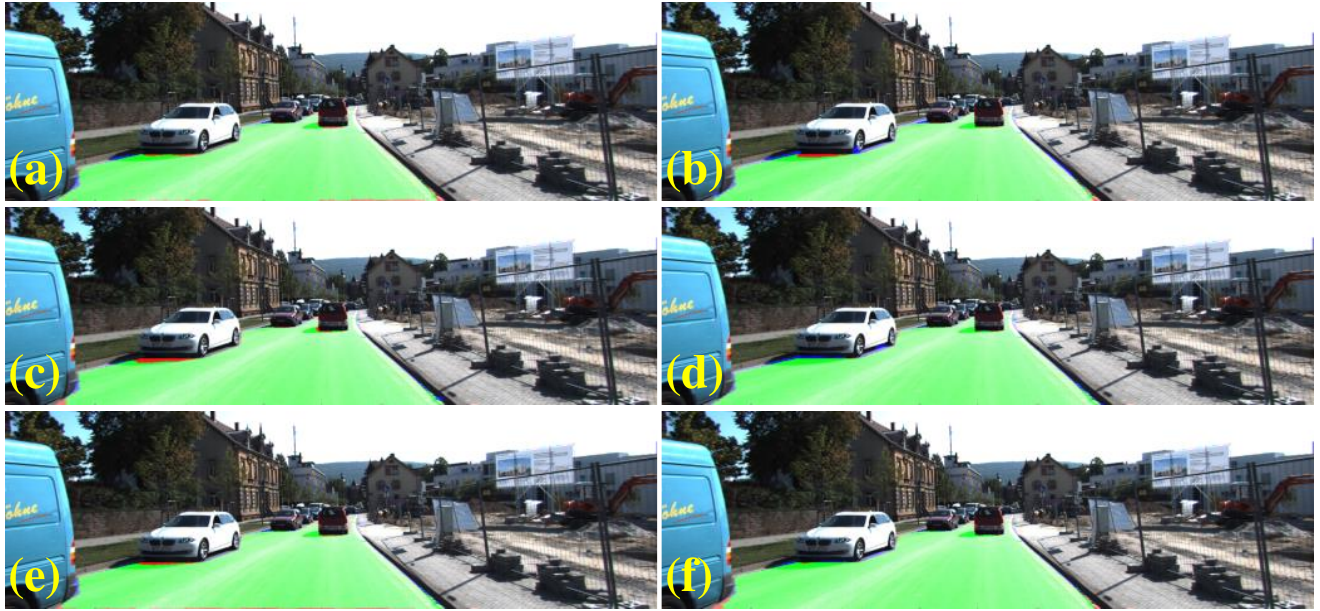}
    \caption{An example of testing images on the KITTI road benchmark, where (a)-(f) shows the road prediction obtained by MultiNet \cite{teichmann2018multinet}, StixelNet II \cite{garnett2017real}, RBNet \cite{chen2017rbnet}, LC-CRF \cite{gu2019road}, LidCamNet \cite{caltagirone2019lidar} and our proposed NIM-RTFNet, respectively. Correctly detected drivable areas are in green. Red pixels correspond to false negatives, whereas blue pixels denote false positives.}
    \label{fig.road}
    \vspace{-1em}
\end{figure}

As previously mentioned, we select our best approach, NIM-RTFNet, and submit its results to the KITTI road benchmark \cite{Fritsch2013ITSC}. The overall performance of our NIM-RTFNet ranks 8th on the KITTI road benchmark. Fig.\ref{fig.road} illustrates an example of KITTI road testing images, and Table \ref{tab.road} presents the evaluation results. We can observe that our proposed NIM-RTFNet outperforms most existing approaches, which confirms the effectiveness and good generalization ability of our proposed NIM. Additionally, although the MaxF of LidCamNet \cite{caltagirone2019lidar} presents slight advantages over ours, our NIM-RTFNet runs much faster than it, and therefore greatly minimizes the trade-off between speed and accuracy.

\begin{table}[t]
    \caption{KITTI road benchmark results, where the best results are in bold type.}
    \centering
    \begin{tabular}{L{2.5cm}C{1.4cm}C{1.4cm}C{1.4cm}}
        \toprule
        Approach & MaxF ($\%$) & AP ($\%$) & Runtime (s) \\ \midrule
        MultiNet \cite{teichmann2018multinet} & 94.88 & 93.71 & 0.17\\
        StixelNet II \cite{garnett2017real} & 94.88 & 87.75 & 1.20 \\
        RBNet \cite{chen2017rbnet} & 94.97 & 91.49 & 0.18\\
        LC-CRF \cite{gu2019road} & 95.68 & 88.34 & 0.18 \\
        LidCamNet \cite{caltagirone2019lidar} & \textbf{96.03} & 93.93 & 0.15 \\ \midrule
        NIM-RTFNet (Ours) & 96.02 & \textbf{94.01} & \textbf{0.05} \\ \bottomrule
    \end{tabular}
    \label{tab.road}
    \vspace{-1em}
\end{table}

\section{Conclusions}
In this paper, we proposed a novel module NIM, which can be easily deployed in various CNNs to refine semantic image segmentation. The experimental results demonstrate that our NIM can greatly enhance the performance of CNNs for the joint detection of drivable areas and road anomalies. Furthermore, our NIM-RTFNet ranks 8th on the KITTI road benchmark and exhibits a real-time inference speed. In the future, we plan to propose a more feasible and computationally efficient cost function for our NIM.

\bibliographystyle{IEEEtran}
\bibliography{egbib}

\end{document}